\documentclass[conference]{IEEEtran}
\IEEEoverridecommandlockouts
\usepackage{hyperref}
\usepackage{amsmath, amssymb, amsfonts, commath}
\usepackage{algorithmic}
\usepackage{graphicx, subfigure}
\usepackage{textcomp}
\usepackage{bm}
\usepackage{xcolor}
\usepackage{multirow}
\usepackage{adjustbox}

\usepackage[
backend=biber,
style=numeric,
sorting=ynt
]{biblatex}
\begin{document}

\title{Inception-Based Crowd Counting --- Being Fast while Remaining Accurate
\thanks{This work was finished under the supervision of Prof. Tanaya Guha, Prof. Victor Sanchez and Prof. Theo Damoulas from the Computer Science Department at the University of Warwick, with the external support from Transport for London.}
}

\author{
\IEEEauthorblockN{Yiming Ma}
\IEEEauthorblockA{\textit{Mathematics for Real-World Systems CDT} \\
\textit{The University of Warwick}\\
Coventry, UK \\
\href{mailto:yiming.ma.1@warwick.ac.uk}{yiming.ma.1@warwick.ac.uk}}}

\maketitle

\begin{abstract}
Recent sophisticated CNN-based algorithms have demonstrated their extraordinary ability to automate counting crowds from images, thanks to their structures which are designed to address the issue of various head scales. However, these complicated architectures also increase computational complexity enormously, making real-time estimation implausible. Thus, in this paper, a new method, based on Inception-V3, is proposed to reduce the amount of computation. This proposed approach (ICC), exploits the first five inception blocks and the contextual module designed in CAN to extract features at different receptive fields, thereby being context-aware. The employment of these two different strategies can also increase the model's robustness. Experiments show that ICC can at best reduce 85.3 percent calculations with 24.4 percent performance loss. This high efficienty contributes significantly to the deployment of crowd counting models in surveillance systems to guard the public safety. The code will be available at https://github.com/YIMINGMA/CrowdCounting-ICC,and its pre-trained weights on the Crowd Counting dataset, which comprises a large variety of scenes from surveillance perspectives, will also open-sourced.
\end{abstract}


\section{Introduction}

\IEEEPARstart{C}{rowd} counting is the computer vision area that concerns estimating the number of individuals present in an image. It has a wide range of applications, such as traffic control \cite{6514618, 7100903, 7351516} and biological studies \cite{lempitsky2010learning, lu2017tasselnet}. One of its most remarkable practices is monitoring the density of people, which is crucial under the circumstance of the global pandemic of COVID-19. Videos from surveillance cameras can be decomposed into frame. Then these frames can be fed into crowd counting models to infer numbers of people. This technique allows governments to know about the crowdedness of a place and take corresponding measures to inhibit the transmission of the virus. Another example is that public transport companies can calculate the number of passengers in a station and adjust the timetable dynamically to reduce the running cost. These two instances demonstrate that crowd counting has a vast potential to be applied in surveillance systems. 

\begin{figure}[htbp]
\centering
\subfigure[A free-view image with annotations from ShanghaiTech A \cite{zhang2016single}.]{
\begin{minipage}[t]{0.45\textwidth}
\centering
\includegraphics[width=\textwidth]{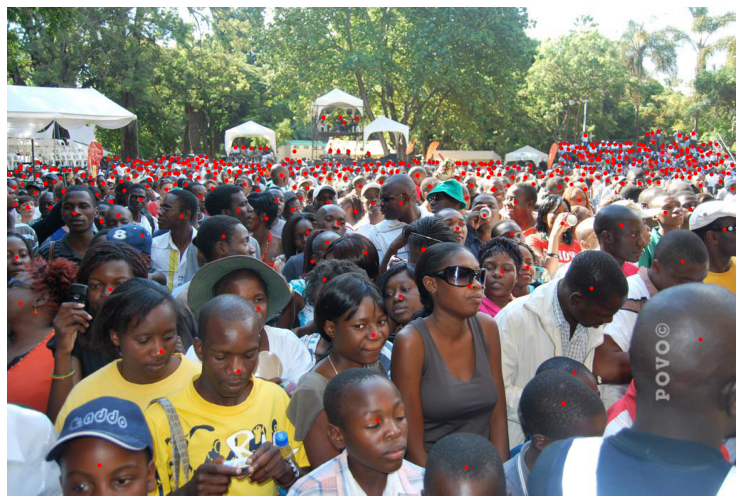}
\end{minipage}
}
\subfigure[A surveillance-view image with annotations from Crowd Surveillance \cite{yan2019perspective}.]{
\begin{minipage}[t]{0.45\textwidth}
\centering
\includegraphics[width=\textwidth]{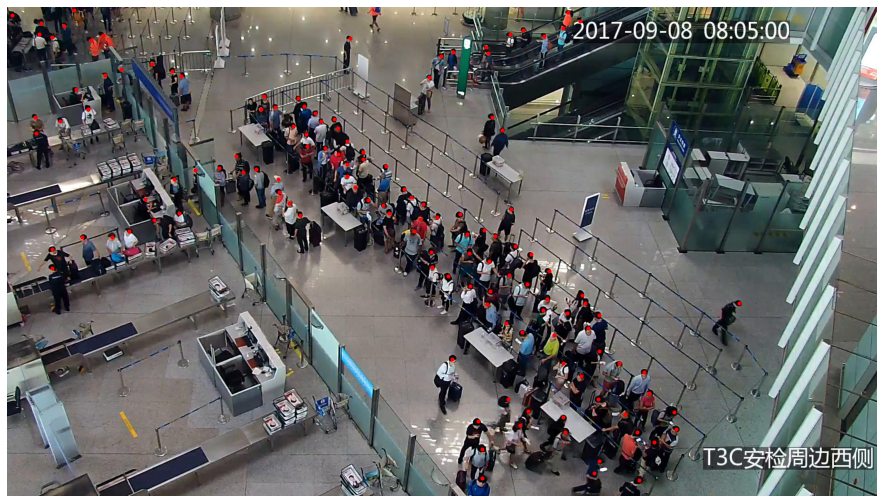}
\end{minipage}
}
\caption{Two instances of images and their corresponding dot annotations, which are usually marked at centers of people's heads. Images from a free view can contain a much more congested crowd than those taken by surveillance cameras.}
\label{fig:1}
\end{figure} 

On the other hand, there are two challenges encountered during the deployment of crowd counting models. Although state-of-the-art algorithms have achieved exceptional results on several benchmark datasets, their backbones slow down the inference speed. Most use VGGs \cite{Simonyan15} as front ends, but this family of neural networks is composed of standard convolutions, indicating that hundreds of billions of arithmetic operations are required to make prediction on a $720$P image. In contrast, many $1\times1$ and spatially separable convolutions are involved in Inception-V3 \cite{szegedy2016rethinking}, significantly reducing the total amount of computation. More detailed comparison of the inference time of different CNN models can be found on \href{https://keras.io/api/applications/}{Keras Applications} \cite{chollet2015keras}.
The other difficulty is the lack of appropriately pre-trained models. The most popular benchmark datasets are UCF\_CC\_50 \cite{6619173}, ShanghaiTech (A and B) \cite{zhang2016single}, and UCF-QNRF \cite{idrees2018composition}, so most researchers only open-source weights tuned on them to prove their models' superiority. However, although ShanghaiTech B \cite{zhang2016single} is from a surveillance perspective, it solely represents crowded scenes, and the other three are free-view and much more congested. Hence, models trained on these datasets usually fail to generalize well to real-world data from surveillance cameras. By comparison, Crowd Surveillance \cite{yan2019perspective} is another surveillance-view dataset containing both sparse and crowded scenes. This property endows models with better generalization on surveillance videos.

Hence, in this paper, inspired by  Inception-V3 \cite{szegedy2016rethinking}, a more diminutive crowd counting model will be proposed, and its pre-trained weights on Crowd Surveillance \cite{yan2019perspective} will be open-sourced to facilitate its implementation in monitoring systems. Besides, multiple experiments will be conducted, including evaluating its performance on ShanghaiTech (A and B) \cite{zhang2016single} and Mall \cite{chen2012feature}, to show that it requires fewer computation resources while preserves a high accuracy.

\section{Related Work}

\subsection{The Evolvement of Crowd Counting Algorithms}

Early crowd counting approaches \cite{983420, 4761705, 5206621} are based on object detection. They require laborious box annotations and are sensitive to occlusion, so later works seek to circumvent detection. Some \cite{5459191, chen2012feature, 6619163} treat crowd counting as a regression problem and directly output the total count from a feature vector. Nevertheless, dot annotations (see Fig. \ref{fig:1}) are underutilized in these algorithms because their loss functions are straight related to the ground-truth total count and its estimation, while the pixel-wise distribution of the crowd, which seems to be more critical, is neglected. As a result, these models suffer from poor generalization, and soon, they have been superseded by algorithms \cite{zhang2015cross, zhang2016single, cao2018scale, liu2018decidenet, lu2018class, ranjan2018iterative, sam2018divide, liu2019recurrent,  liu2019adcrowdnet, liu2019context, shi2019revisiting, wan2019adaptive, xiong2019open, zhang2019relational, wang2020distribution, song2021choose, wan2021generalized} that instead predict the crowd density, whose values indicate the number of people in the corresponding pixels.

However, the development of density-prediction-based approaches is not smooth; the problem of variant local scales of heads influencing a model's performance has been haunting researchers for years. One viable solution is to leverage geometric information \cite{zhang2015cross, kang2017incorporating} to adjust models to the scene's geometry, but this information is usually unavailable in the test environment. Thus, other methods choose to include modules designed to cope with the variation of local scales. One of the most prominent among them is CAN \cite{liu2019context},  in which rapid scale changes are being handled by a contextual-aware module which fuses multi-scale features adaptively. Nevertheless, its backbone is still based on VGG-16 \cite{Simonyan15}, resulting in extensive computation and thus improper for real-time estimation. 

\subsection{Inception-V3}

In comparison, Inception-V3 \cite{szegedy2016rethinking} is a more suitable candidate for the front end of fast algorithms since it has much fewer standard convolutions by decomposing kernels. For example, a convolutional kernel with a filter size of $3\times3$ and $64$ output channels can be replaced with a $1\times1$ kernel that add values along the channel, followed by a $3\times3$ kernel to increase the number of channels. The former one performed on a $4\times4$ image would require $13,568$ arithmetic operations, while the latter only needs $4,432$, reducing around $67\%$ computation. Large kernels ($n \times n$) can also be factorized into a $n \times 1$ kernel and a $1 \times n$ kernel. Furthermore, unlike VGGs, Inceptions \cite{szegedy2015going, szegedy2016rethinking, szegedy2017inception} have multi-column modules, so they can natively capture the scale variance. Nevertheless, it is insufficient for these modules alone to solve the problem caused by scales.

\subsection{The DM-Count Loss}

Loss functions for training, defined on density maps, have also been well studied. Since the ground-truth density map is a sparse binary matrix, with $0$s and $1$s exceptionally unevenly distributed, functions directly based on the it are difficult to train. Thus, early methods \cite{idrees2013multi, zhang2016single, idrees2018composition, liu2019context, wan2019adaptive} use Gaussian kernels to smooth the binary matrix, and as a result, the model's performance is heavily affected by the quality of smoothing. However, setting suitable kernel widths is not easy because of various scales in an image. Therefore, the DM-Count loss, which uses optimal transport to measure the distance between density map distributions, has been proposed in \cite{wang2020distribution} to address the challenge brought by training and Gaussian smoothing. The authors of \cite{wang2020distribution} have also proved that models trained under the supervision of the DM-Count loss will have tighter upper bounds for the generalization error.

In this work, the first several blocks of Inception-V3 \cite{szegedy2016rethinking} will be used to construct the front end to speed up inference. The advantage of the context-aware module within CAN \cite{liu2019context} will also be leveraged to adapt the model to different head scales. The inception blocks will also be exploited to reinforce the capture of contextual information. The DM-Count loss \cite{wang2020distribution} will be employed in training to avoid issues caused by Gaussian smoothing.

\section{Inception-Based Crowd Counting}

\subsection{Problem Formulation}

Let $\bm{X} \in \bm{\mathcal{X}} \subset \mathbb{R}_+^{h \times w \times 3}$ be an RGB image with the height $h$ and the width $w$ and $\bm{Y} \in \bm{\mathcal{Y}} \subset \mathbb{R}_+^{h \times w}$ be its corresponding binary density map, i.e., for a pixel $(r, c)$, $\bm{Y}_{r, c} = 1$ if and only if there is a person's signiture (usually the head's center) present in it. Assume $p(\bm{X})$ is the distribution of $\bm{X}$ and the dependent variable $\bm{Y}$ is always observable. For a general crowd counting algorithm and a given loss function $\mathcal{L}$ defined on $\bm{\mathcal{Y}} \times \bm{\mathcal{Y}}$, the mathematical aim is to fit a function $\bm{F}(\cdot\, ; \, \bm{\omega})$ parameterized by $\bm{\omega}$ that minimizes the expected cost:
\begin{equation*}
    \int_{\bm{\mathcal{X}}} \mathcal{L} \left( \bm{F}(\bm{X}; \, \bm{\omega}), \, \bm{Y} \right) \dif \bm{X}.
\end{equation*}
Denote its minimum value as $\mathcal{C}_{\bm{F}}$, i.e. 
\begin{equation}
    \mathcal{C}_{\bm{F}} := \min_{\bm{\omega}} \int_{\bm{\mathcal{X}}} \mathcal{L} \left( \bm{F}(\bm{X}; \, \bm{\omega}), \, \bm{Y} \right) \dif \bm{X}.
\end{equation}

However, in practice, the distribution of $\bm{X}$ is unknow, and only observations of $\bm{X}$ and $\bm{Y}$ are accessible, so the problem of overfitting is inevitable. Thus, in machine learning, these data are usually split into two disjoint sets, namely, a training set $(\bm{\mathcal{X}}_\text{tr}, \bm{\mathcal{Y}}_\text{tr})$ and a test set $(\bm{\mathcal{X}}_\text{t}, \bm{\mathcal{Y}}_\text{t})$. The former is usually used to obtain $\hat{\bm{\omega}}$ by optmizing the training cost
\begin{equation*}
    \sum_{\bm{X} \in \bm{\mathcal{X}}_\text{tr}} \mathcal{L}_\text{tr} \left( \bm{F}(\bm{X}; \, \bm{\omega}), \, \bm{Y} \right),
\end{equation*}
while the latter is used to estimate $\mathcal{C}_{\bm{F}}$ by calculating
\begin{equation*}
    \tilde{\mathcal{C}}_{\bm{F}} := \sum_{\bm{X} \in \bm{\mathcal{X}}_\text{t}} \mathcal{L} \left( \bm{F}(\bm{X}; \, \hat{\bm{\omega}}), \, \bm{Y} \right).
\end{equation*}
Since this paper is interested in real-time estimation, the form of $\bm{F}$ is constrained within a particular range $\bm{\mathcal{F}}$. The prediction must also remain as precise as possible, so ideally, the ultimate purpose is to solve
\begin{equation} \label{eqn1}
    \min_{\bm{F} \in \bm{\mathcal{F}}} \tilde{C}_{\bm{F}}.
\end{equation}

\begin{figure*}[hbtp]
    \centering
    \includegraphics[width=\textwidth]{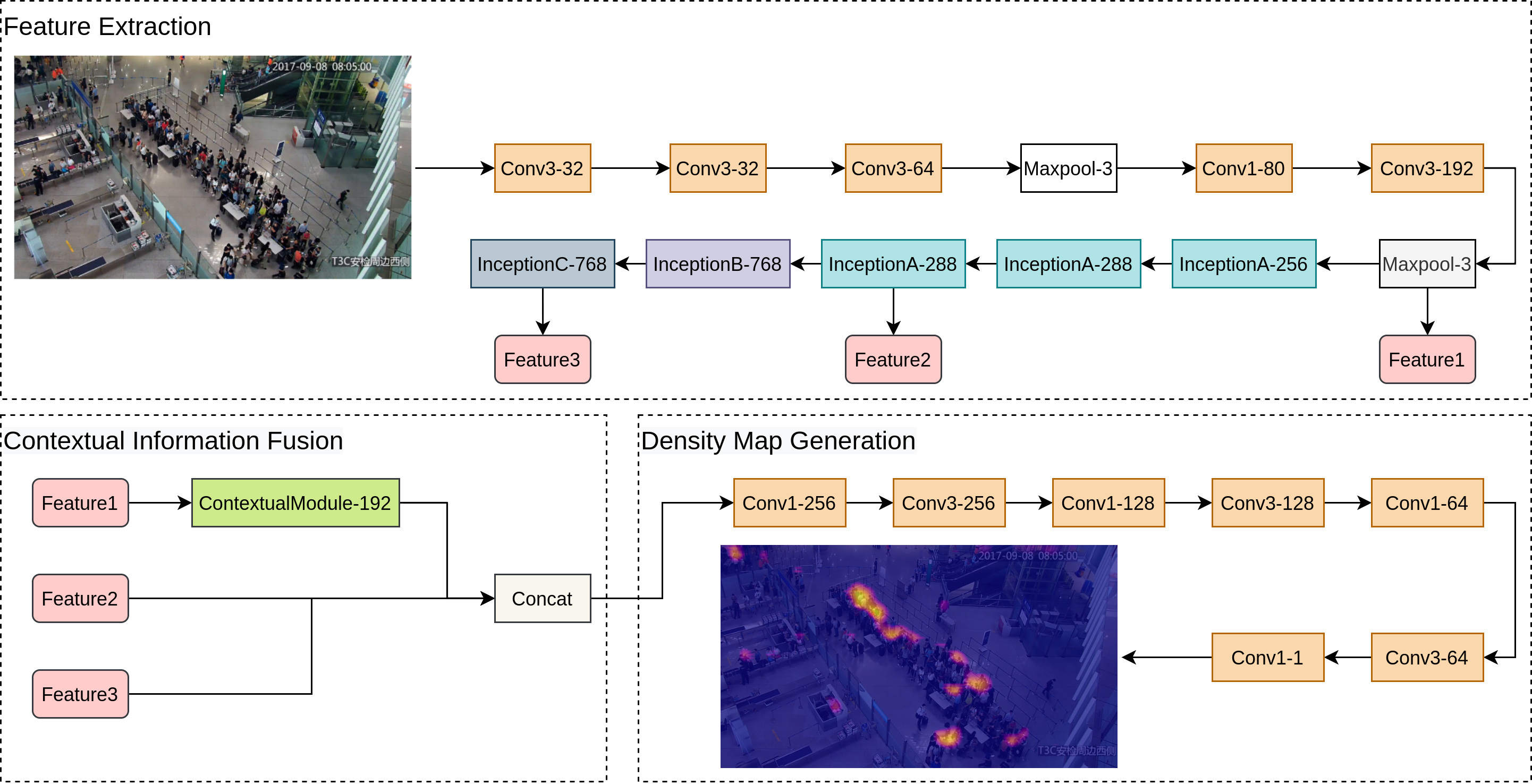}
    \caption{The architecture of the proposed model. It uses the first twelve layers of Inception-V3 \cite{szegedy2016rethinking} to obtain features at different levels from the input image. The contextual module from CAN \cite{liu2019context} will be leverage to process \texttt{Feature1}. Then they are fused and passed into the decoder to generate the predicted density map.}
    \label{fig:2}
\end{figure*}

\subsection{Model Architecture}

Due to the unfeasibility to solve \eqref{eqn1}, and considering the benefits of those models mentioned in Section II, this paper comes up with a suboptimal solution instead. Its structure is shown in Fig. \ref{fig:2}.

\subsubsection{The Encoding Front End}

The starting point is a network consisting of the first 12 blocks of a pre-trained Inception-V3 \cite{szegedy2016rethinking}. \texttt{Feature1}, which is the output from the second max-pooling layer, is analogous to the base features extracted by the VGG component in CAN \cite{liu2019context}. The difference is that \texttt{Feature1} comes from a much shallower layer, accommodating more rich details. As the it goes deeper, the output from the third Inception-A block \cite{szegedy2016rethinking} is cloned as \texttt{Feature2}. To avoid introducing too many parameters, the original Inception-V3 \cite{szegedy2016rethinking} is truncated after the first Inception-C block \cite{szegedy2016rethinking}, whose output is denoted as \texttt{Feature3}.

\subsubsection{Contextual Information Extraction}

There are two different approaches used to extract contextual information. One is to impose the contextual module from \cite{liu2019context}, in which filters with different sizes (1, 2, 3 and 6) are employed, on \texttt{Feature1} to produce scale-aware features. As the $1 \times 1$ kernel is utilized, abundant details can be well preserved. These processed features are also context-aware, contributing remarkably to accurate predictions of heads at both small and large scales. The other avenue is to utilize the Inception blocks \cite{szegedy2016rethinking} within the front end. These blocks also have bottleneck structures to learn both sparse and non-sparse features, and the dissimilarity is that they are much deeper and during the extraction of \texttt{Feature2} and \texttt{Feature3}, contextual information is repeatedly fused. As \texttt{Feature2} and \texttt{Feature3} already contains contextual information, there is no further processing of them, and all these context-aware features are finally combined along the channel. Exploiting disparate strategies can reinforce the model's robustness.

\subsubsection{The Decoding Back End}

The concatenated features are being input into this component to generate the predicted density map. $1\times1$ convolutions are executed before standard convolutions to decrease the number of channels, so the overall computation can be notably reduced. Finally, tensor values are summed up across the channel to generate the predicted density map. Note that since downsampling operations like pooling with the stride of 2 are used, the size of the predicted density map is 8 times smaller. To recover it to the full size, upsampling such as interpolation can be leveraged.

\subsection{Training Details}

\subsubsection{Pre-Processing}

As for pre-processing, because of the reduced output size, the ground-truth density map needs to be downsampled to have the same shape. Besides, since widths and heights of images usually differ within any dataset, cropping is utilized so that data can be batched in training.

\subsubsection{Loss Functions}

For the processed ground truth $\bm{Y}$ and its approximation $\hat{\bm{Y}}$, the loss function for training is the DM-Count loss defined in \cite{wang2020distribution}, which comprises the counting loss $\ell_\text{C}$, the optimal transport loss $\ell_\text{OT}$ and the total variation loss $\ell_\text{TV}$.

As the optimal transport measures the dissimilarity between two probability distributions, vectorizing $\bm{Y}$ and $\hat{\bm{Y}}$ brings easier definitions. Denote the downsampled width and height as $w'$ and $h'$. Let $\bm{y}$ and $\hat{\bm{y}}$ be the unrolled forms of $\bm{Y}$ and $\hat{\bm{Y}}$, so $\bm{y}, \, \hat{\bm{y}} \in \mathbb{R}^{h' \times w'}$. Since each element of $\bm{y}$ (or $\hat{\bm{y}}$) represents the corresponding crowd density at that pixel, the total count is exactly its $L_1$ norm. Thus, the counting loss is defined as
\begin{equation}
    \ell_\text{C}(\bm{y}, \, \hat{\bm{y}}) := \left| \| \bm{y} \|_1 - \| \hat{\bm{y}} \|_1 \right|,
\end{equation}
which measures the difference of the ground-truth and predicted counts.

Similar to the Kullback-Leibler divergence and the Jensen-Shannon divergece, Monge-Kantorovich's optimal transport cost is another way to measure the distance between two probability distributions, but it has certain better optimization properties \cite{arjovsky2017wasserstein} and therefore is exploited. Assume $\bm{p}$ and $\bm{q}$ are two probability masses, i.e., $\bm{p}_i, \, \bm{q}_i \ge 0$ and $\sum_{i = 1}^n \bm{p}_i = \sum_{j = 1}^n \bm{q}_j = 1$. Let $c(\bm{p}_i, \, \bm{q}_j)$ be the cost of transforming $\bm{p}_i$ to $\bm{q}_j$ and $\bm{C} \in \mathbb{R}_+^{n \times n}$ such that $\bm{C}_{i,j} = c(\bm{p}_i, \, \bm{q}_j)$. Then the optimal transport cost $\mathcal{W}(\bm{p}, \, \bm{q})$ is defined as the minimum cost of transforming $\bm{p}$ into $\bm{q}$, which is
\begin{equation} \label{eqn2}
    \mathcal{W}(\bm{p}, \, \bm{q}) := \min_{\bm{\pi} \in \mathbb{R}^{n \times n}_+} \langle \bm{\pi}, \, \bm{C} \rangle.
\end{equation}
For $\bm{\pi} \in \mathbb{R}^{n \times n}_+$ to be a valid transformation, it needs to satisfy $\sum_{k = 1}^n \bm{\pi}_{i, k} = \bm{p}_i$ and $\sum_{k = 1}^n \bm{\pi}_{k, j} = \bm{q}_j$, $\forall i, j \in \mathbb{N}_+$, $1 \le i, \, j \le n$.

Hence, in \cite{wang2020distribution}, by viewing $\bm{y}$ and $\hat{\bm{y}}$ as two unnormalized distributions, the optimal transport loss is defined as 
\begin{equation}
    \ell_\text{OT} (\bm{y}, \, \hat{\bm{y}}) := \mathcal{W} \left( \frac{\bm{y}}{\| \bm{y} \|_1}, \, \frac{\hat{\bm{y}}}{\| \hat{\bm{y}} \|_1} \right),
\end{equation}
in which the transport cost function is defined as the squared Euclidean distance of corresponding pixels. For example, suppose $\bm{y}_{h}$ corresponds to pixel $(i, \, j)$ in $\bm{Y}$ and $\hat{\bm{y}}_k$ corresponds to pixel $(p, \, q)$ in $\hat{\bm{Y}}$, then
\begin{equation}
    c \left( \frac{\bm{y}_h}{\| \bm{y} \|_1}, \, \frac{\hat{\bm{y}}_k}{\| \hat{\bm{y}} \|_1} \right) := ( p - i )^2 + (q - j)^2.
\end{equation}

Notice that the definition of \eqref{eqn2} involves an optimization problem. Different from \cite{wang2020distribution}, in this paper, it is solved by the Sinkhorn-Knopp matrix scaling algorithm proposed in \cite{cuturi2013sinkhorn} and implemented by POT \cite{flamary2021pot}. Since numerical errors will also be introduced in this process, the total variation loss defined in \cite{wang2020distribution} is still included to stablize training:
\begin{equation}
    \ell_\text{TV} (\bm{y}, \, \hat{\bm{y}}) := \frac{1}{2} \left\| \frac{\bm{y}}{\| \bm{y} \|_1} - \frac{\hat{\bm{y}}}{\| \hat{\bm{y}} \|_1} \right\|_1.
\end{equation}
The overall loss function \cite{wang2020distribution} is defined as
\begin{equation}
    \ell (\bm{y}, \, \hat{\bm{y}}) := \ell_\text{C}(\bm{y}, \, \hat{\bm{y}}) + \lambda_1 \ell_\text{OT} (\bm{y}, \, \hat{\bm{y}}) + \lambda_2  \| \bm{y} \|_1 \ell_\text{TV} (\bm{y}, \, \hat{\bm{y}}),
\end{equation}
in which $\lambda_1$ and $\lambda_2$ are tunable hyper-parameters.

\subsubsection{Optimization}

AdamW \cite{loshchilov2018decoupled} with exponential learning rate decay is used as the optmizer, and the initial learning rate used on all datasets is $10^{-4}$. The model's performance on validation sets is evaluated after each epoch to avoid overfitting to training sets.

\section{Experiments}

In this section, the proposed model (ICC) will be evaluated, so evaluation metrics and benchmark datasets will be first introduced. Then it will be compared with the state-of-the-art approaches, and finally, an ablation study about the two avenues to extract contextual information will be conducted.

\subsection{Evaluation Metrics}

Following previous works \cite{zhang2016single, babu2017switching, sindagi2017cnn, li2018csrnet, liu2019context, wang2020distribution, thanasutives2021encoder, wan2021generalized}, the mean absolute error (MAE) and the root mean squared error (RMSE) are utilized as two evaluation metrics. For a set of ground-truth counts $\{ {z}_1, \, \cdots, \, {z}_N \}$ and their predictions $\{ \hat{{z}}_1, \cdots, \, \hat{{z}}_N \}$, these two metrics are defined as
\begin{equation}
    \text{MAE} := \frac{1}{N} \sum_{i=1}^N | z_i - \hat{z}_i |,
\end{equation}
and
\begin{equation}
    \text{RMSE} := \sqrt{\frac{1}{N} \sum_{i=1}^N (z_i - \hat{z}_i )^2}.
\end{equation}
The number of floating-point operations required to infer on a $1080$P ($1080 \times 1920$) image is leveraged to quantify the each model's computational complexity.

\subsection{Benchmark Datasets}

Three datasets, ShanghaiTech A \cite{zhang2016single}, ShanghaiTech B \cite{zhang2016single} and Mall \cite{chen2012feature} are used for comparison, although there are some other classic benchmark datasets in crowd counting, such as UCF\_CC\_50 \cite{idrees2013multi} and UCF-QNRF \cite{idrees2018composition}. This is because the primary purpose of this paper is to provide an efficient model for surveillance systems, only one free-view dataset (which is ShanghaiTech A \cite{zhang2016single}) is needed to reflect the model's general crowd counting capability. The Crowd Surveillance dataset \cite{yan2019perspective} will also be introduced, as the power of weights trained on it is demonstrated in Section \ref{sec:model_comparison}.

\subsubsection{\texorpdfstring{ShanghaiTech A \cite{zhang2016single}}{ShanghaiTech A}}

\begin{figure}[htbp]
\centering
\subfigure[]{
\centering
\includegraphics[width=0.145\textwidth]{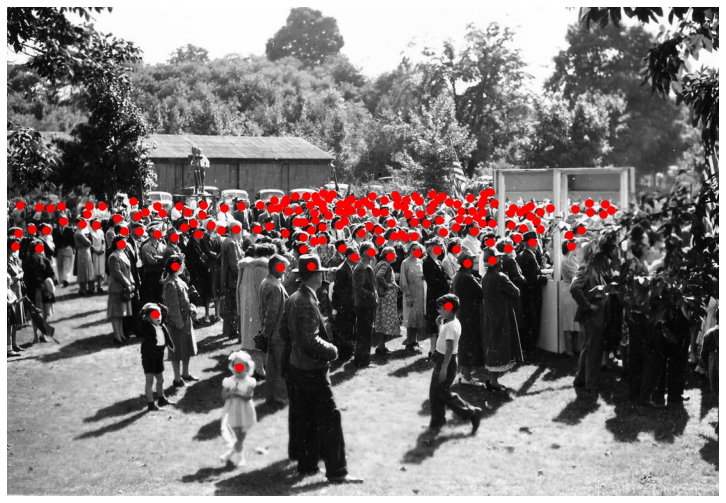}
\label{fig:3.1}
}
\subfigure[]{
\centering
\includegraphics[width=0.145\textwidth]{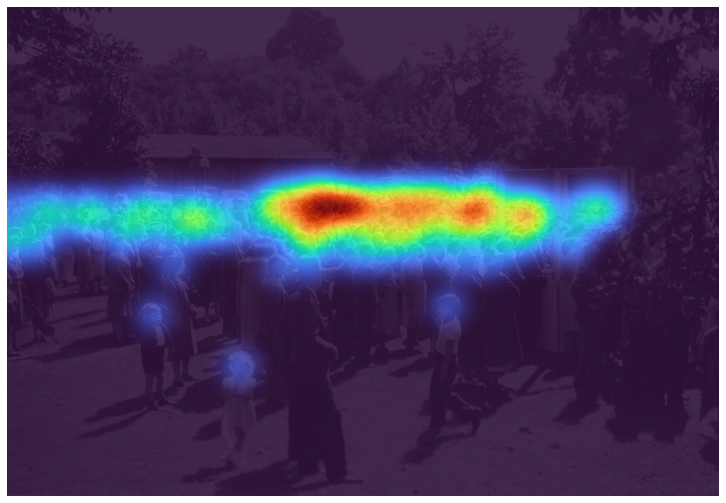}
\label{fig:3.2}
}
\subfigure[]{
\centering
\includegraphics[width=0.145\textwidth]{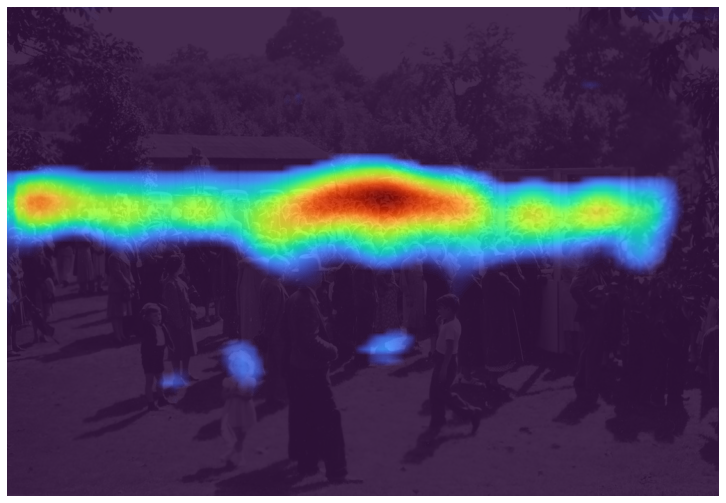}
\label{fig:3.3}
}
\subfigure[]{
\centering
\includegraphics[width=0.145\textwidth]{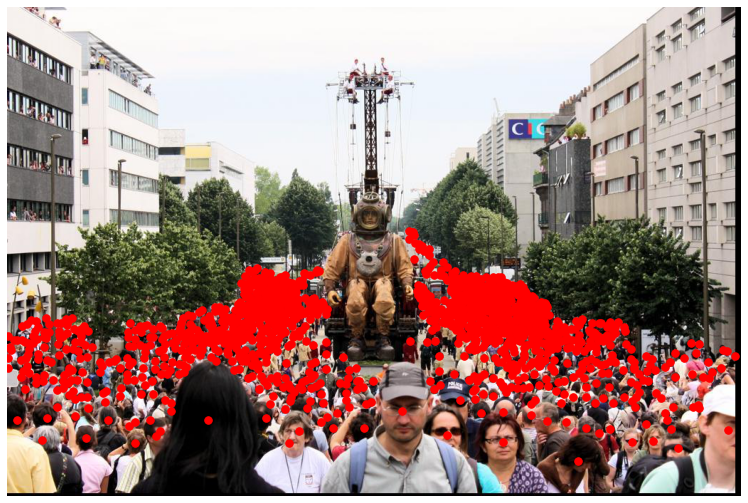}
\label{fig:3.4}
}
\subfigure[]{
\centering
\includegraphics[width=0.145\textwidth]{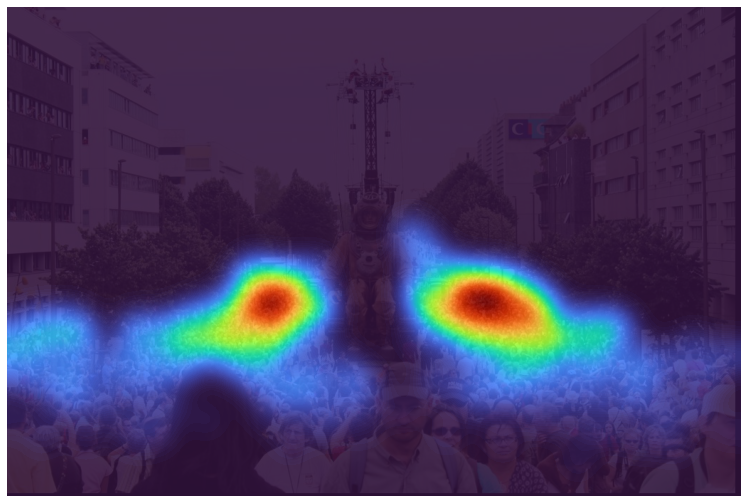}
\label{fig:3.5}
}
\subfigure[]{
\centering
\includegraphics[width=0.145\textwidth]{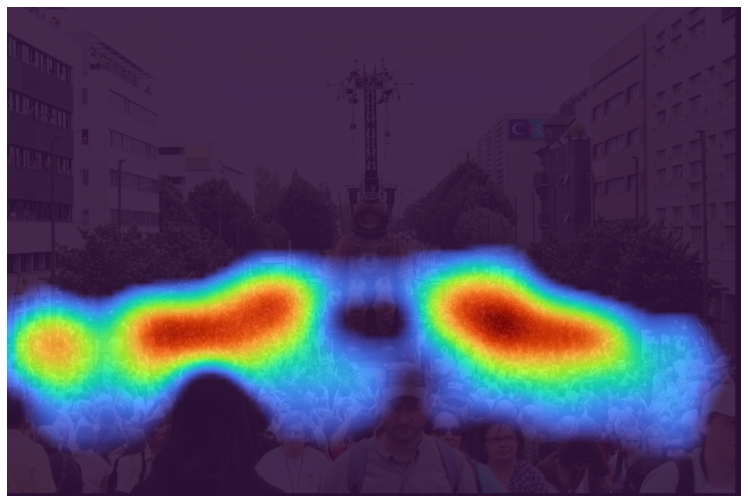}
\label{fig:3.6}
}
\caption{Crowd density estimation on two images from ShanghaiTech A \cite{zhang2016single}. \ref{fig:3.1} and \ref{fig:3.4} are images annotated with original ground truths, whose corresponding Gaussian smoothed (with $\sigma=20$) alternates are shown in \ref{fig:3.2} and \ref{fig:3.5}. The proposed algorithm's predictions are visualized in \ref{fig:3.3} and \ref{fig:3.6}, which demonstrate its good performance on unseen data.}
\label{fig:3}
\end{figure}

It is composed of $482$ images with annotations, $300$ for training and $182$ for testing. The average resolution of these images is $589 \times 868$, and the average number of individuals present in one is approximately $501$. These images are not guaranteed to be from surveillance cameras, so this dataset is only used to prove the model's ability to count crowds in free-view images. The orignal training set is split into a new training set ($80\%$) and a validation set ($20\%$). The latter is designated to tune hyper-paramters, such as the number of epochs for training. Fig. \ref{fig:3} provides visualization of the proposed model's predictions on two randomly selected test images from ShanghaiTech A \cite{zhang2016single}.

\subsubsection{\texorpdfstring{ShanghaiTech B \cite{zhang2016single}}{ShanghaiTech B}}

\begin{figure}[htbp]
\centering
\subfigure[]{
\centering
\includegraphics[width=0.145\textwidth]{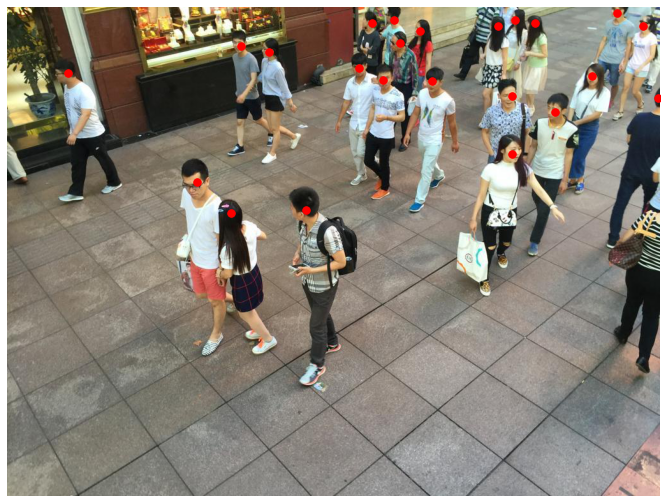}
\label{fig:4.1}
}
\subfigure[]{
\centering
\includegraphics[width=0.145\textwidth]{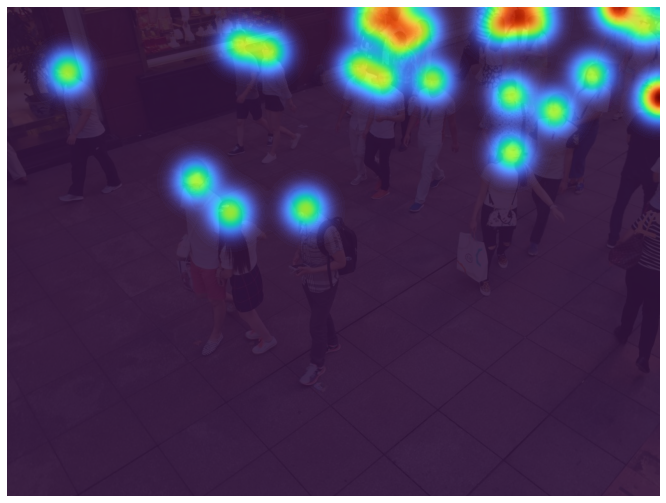}
\label{fig:4.2}
}
\subfigure[]{
\centering
\includegraphics[width=0.145\textwidth]{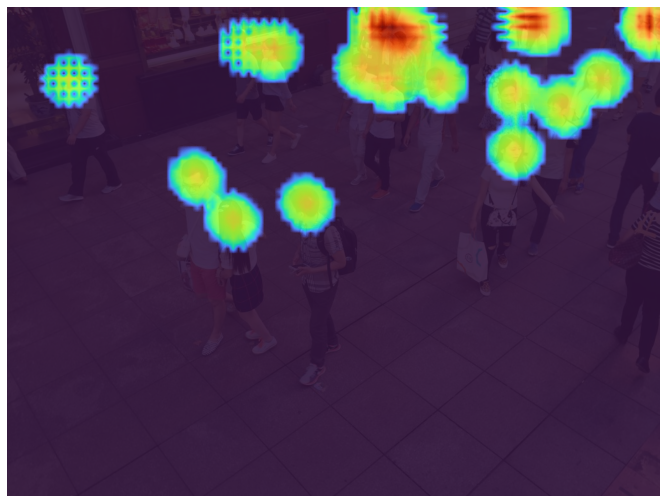}
\label{fig:4.3}
}
\subfigure[]{
\centering
\includegraphics[width=0.145\textwidth]{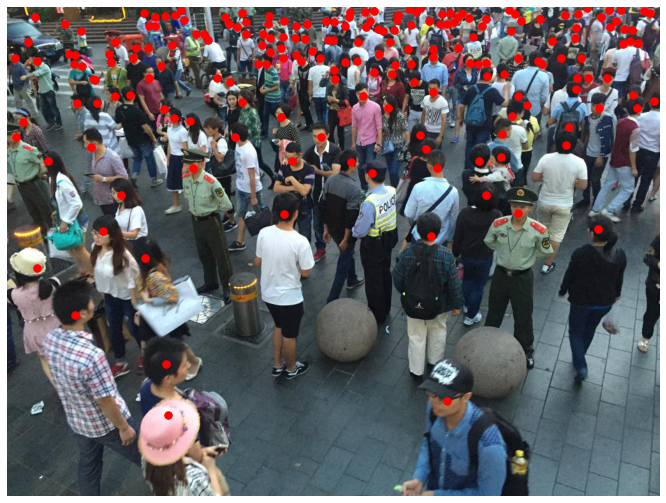}
\label{fig:4.4}
}
\subfigure[]{
\centering
\includegraphics[width=0.145\textwidth]{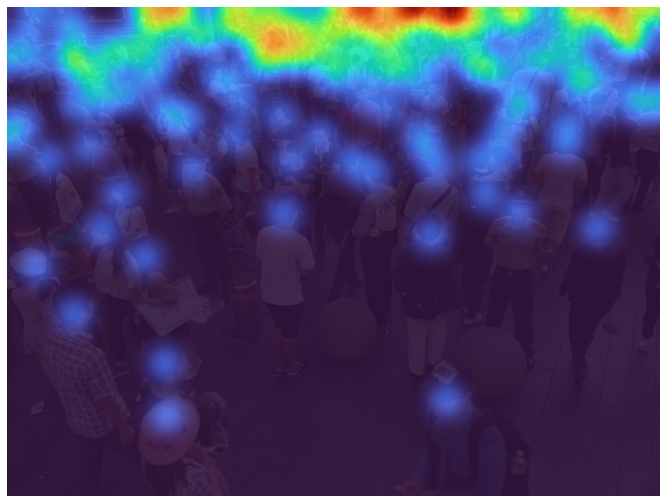}
\label{fig:4.5}
}
\subfigure[]{
\centering
\includegraphics[width=0.145\textwidth]{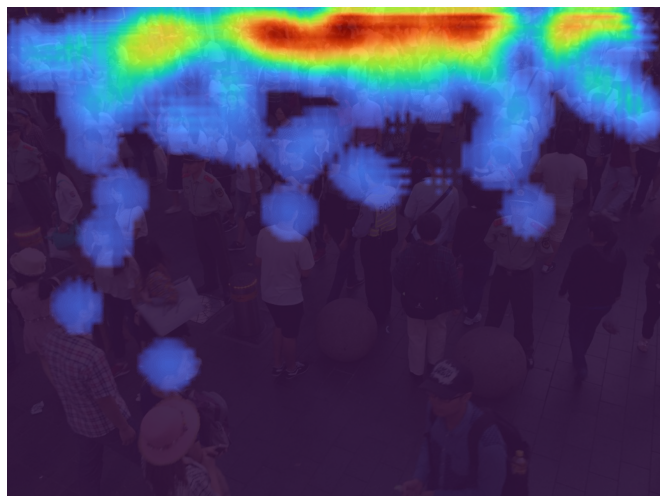}
\label{fig:4.6}
}
\caption{\ref{fig:4.1} and \ref{fig:4.4} are two test images from ShanghaiTech B \cite{zhang2016single}, annotated with original ground truths. The Gaussian kernel with $\sigma=20$ is adopted in smoothing to produce \ref{fig:4.2} and \ref{fig:4.5}, and \ref{fig:4.3} and \ref{fig:4.6} are plots of the corresponding predictions.}
\label{fig:4}
\end{figure}

This dataset comprehends $400$ images for training and $316$ for testing. These images come from surveillance cameras in different scenes. The average resolution is $768 \times 1024$ and the mean count is about $123$, which indicates that crowds in these images are densely distributed. The same data splitting strategy is exploited to generate a validation dataset, and the proposed approach's good generalization is illustrated in Fig. \ref{fig:4}.

\subsubsection{\texorpdfstring{Mall \cite{chen2012feature}}{Mall}}

\begin{figure}[htbp]
\centering
\subfigure[]{
\centering
\includegraphics[width=0.145\textwidth]{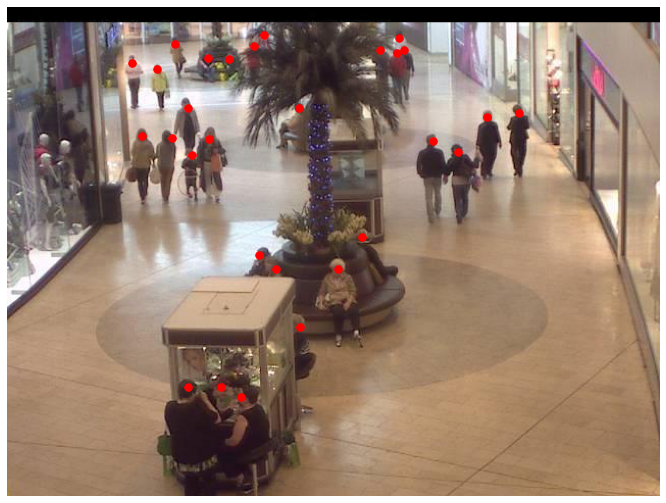}
\label{fig:6.1}
}
\subfigure[]{
\centering
\includegraphics[width=0.145\textwidth]{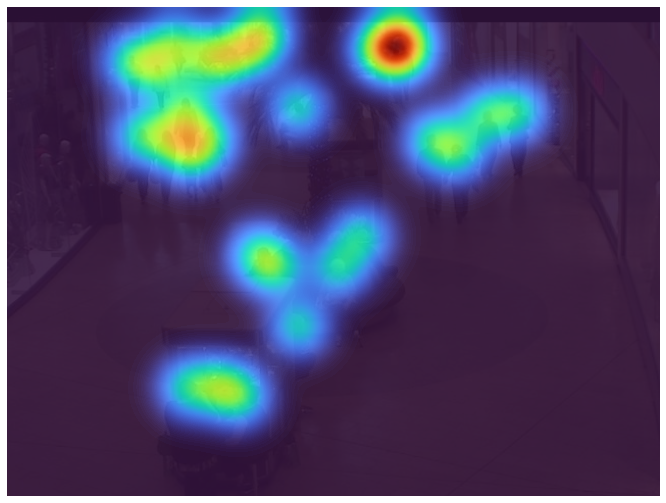}
\label{fig:6.2}
}
\subfigure[]{
\centering
\includegraphics[width=0.145\textwidth]{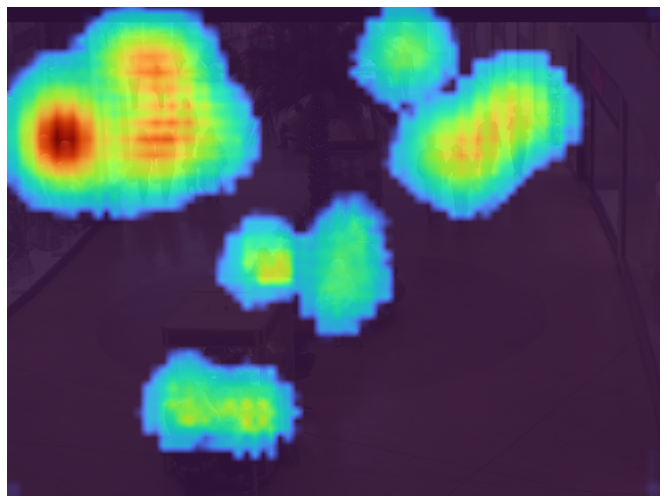}
\label{fig:6.3}
}
\subfigure[]{
\centering
\includegraphics[width=0.145\textwidth]{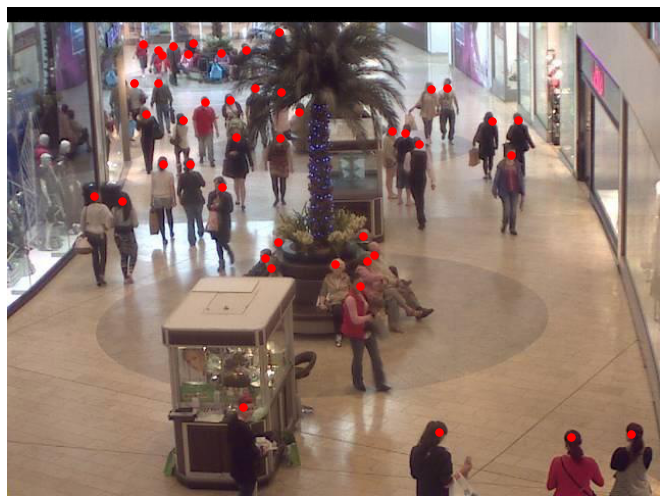}
\label{fig:6.4}
}
\subfigure[]{
\centering
\includegraphics[width=0.145\textwidth]{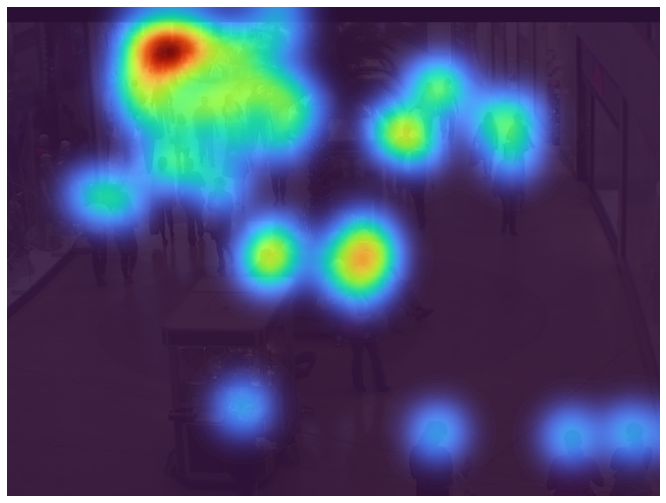}
\label{fig:6.5}
}
\subfigure[]{
\centering
\includegraphics[width=0.145\textwidth]{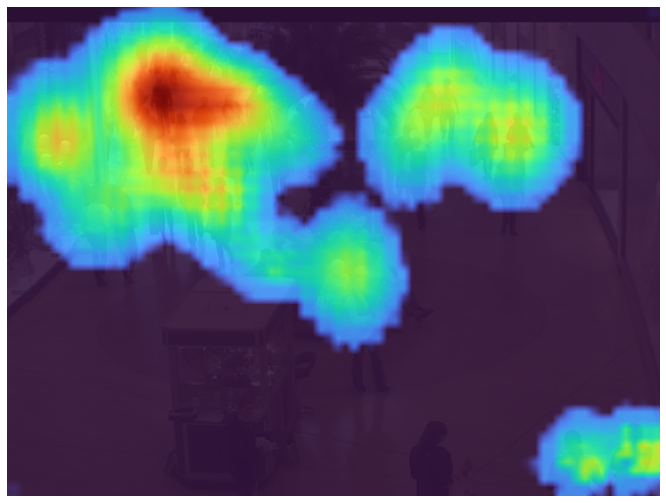}
\label{fig:6.6}
}
\caption{Crowd density predictions of two test images from Mall \cite{chen2012feature}.}
\label{fig:6}
\end{figure}

Although it is also surveillance-view, its difference from ShanghaiTech B \cite{zhang2016single} is that its images are frames of a surveillance video, whose source is a fixed camera at a shopping center. These images are all $480 \times 640$ with the mean count of $31$. Since the principal goal of the proposed method is to become applicable in surveillance systems, this dataset is the most ideal one for evaluation. Following previous works \cite{chen2012feature, xiong2017spatiotemporal, liu2018decidenet, liu2018crowd, hossain2019crowd, zhou2021locality}, the training set comprises the first $800$ frames, and test set is composed of the rest. Results of predictions of two test images are shown in Fig. \ref{fig:6}.

\subsubsection{\texorpdfstring{Crowd Surveillance \cite{yan2019perspective}}{Crowd Surveillance}}

\begin{figure}[htbp]
    \centering
    \includegraphics[width=0.48\textwidth]{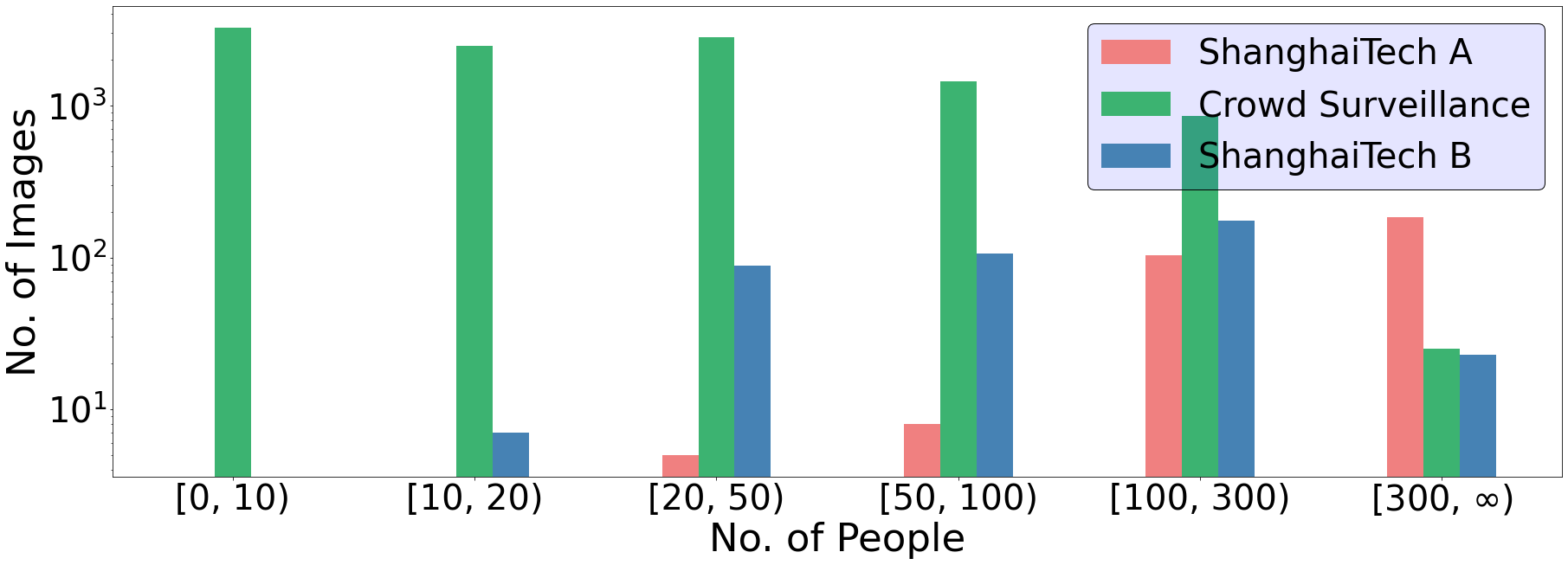}
    \caption{A histogram of total counts from the training sets. As it presents, the total counts of crowds in Crowd Surveillance is more uniformly distributed, so models trained on it should possess excellent generalization on others.}
    \label{fig:5}
\end{figure}

It is one of the largest published crowd counting datasets that are from surveillance viewpoints: $10,880$ images constitute the training set, and $3,065$ account for the test set. More crucially, these images are from numerous free scenes instead of those fixed, and the total count spans a large range, so models trained on it can be exposed to more various situations and thus become more robust. Fig. \ref{fig:5} shows comparison of the distribution of total counts of Crowd Surveillance \cite{yan2019perspective} against those of ShanghaiTech (A and B) \cite{zhang2016single}.

\begin{figure}[htbp]
\centering
\subfigure[]{
\centering
\includegraphics[width=0.145\textwidth]{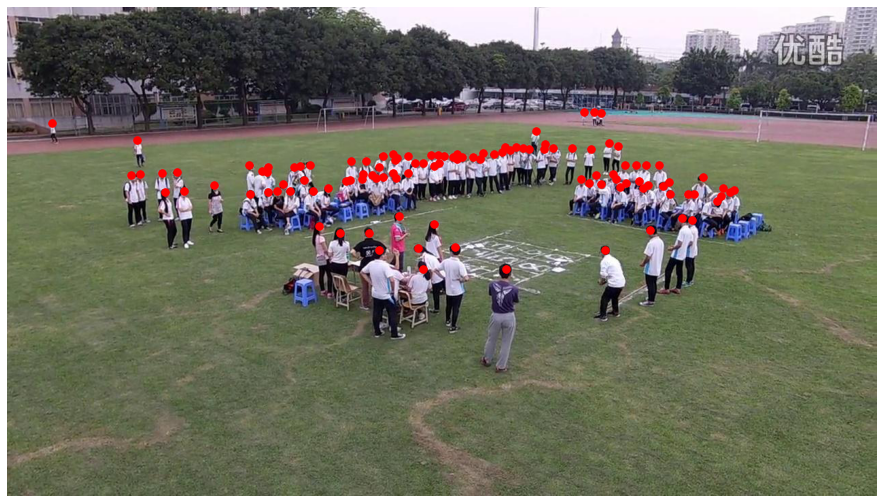}
\label{fig:7.1}
}
\subfigure[]{
\centering
\includegraphics[width=0.145\textwidth]{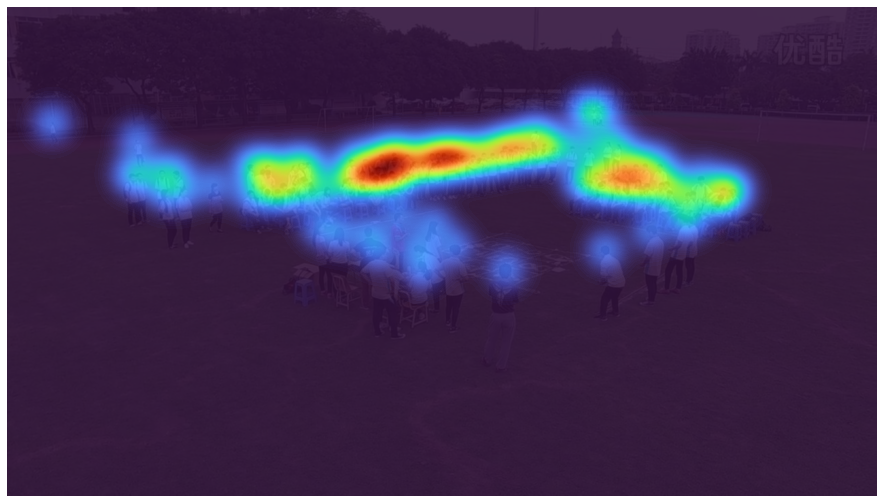}
\label{fig:7.2}
}
\subfigure[]{
\centering
\includegraphics[width=0.145\textwidth]{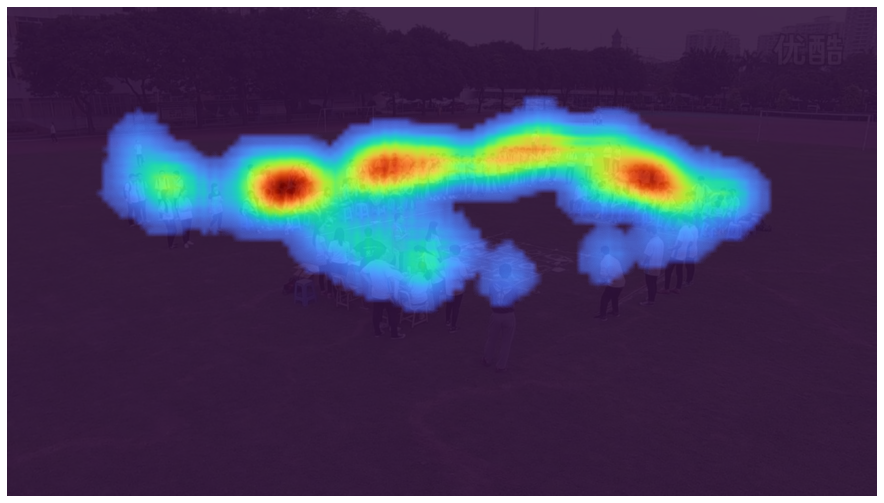}
\label{fig:7.3}
}
\subfigure[]{
\centering
\includegraphics[width=0.145\textwidth]{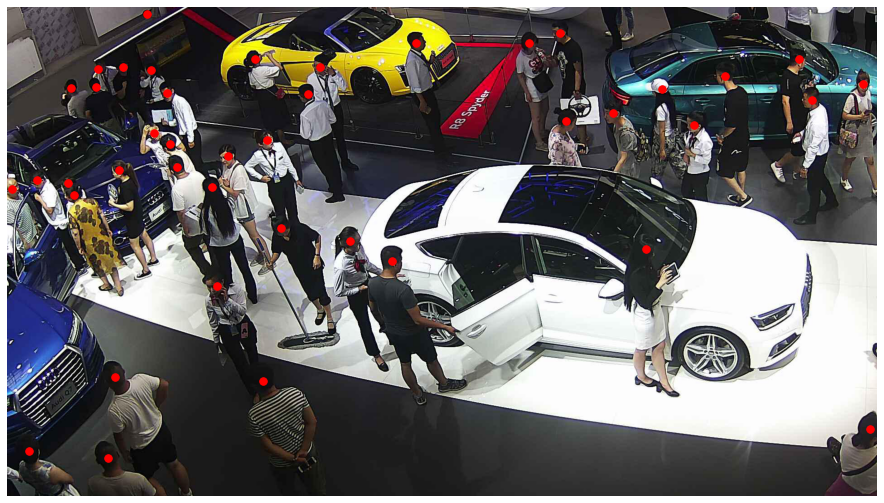}
\label{fig:7.4}
}
\subfigure[]{
\centering
\includegraphics[width=0.145\textwidth]{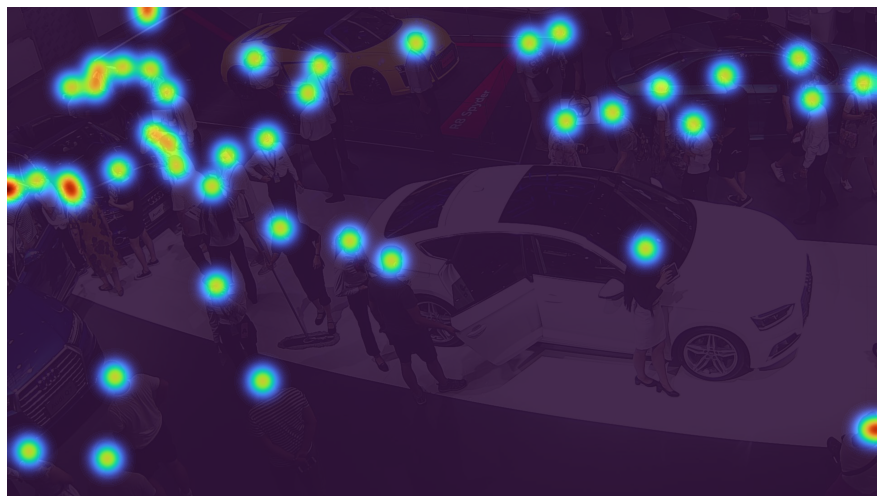}
\label{fig:7.5}
}
\subfigure[]{
\centering
\includegraphics[width=0.145\textwidth]{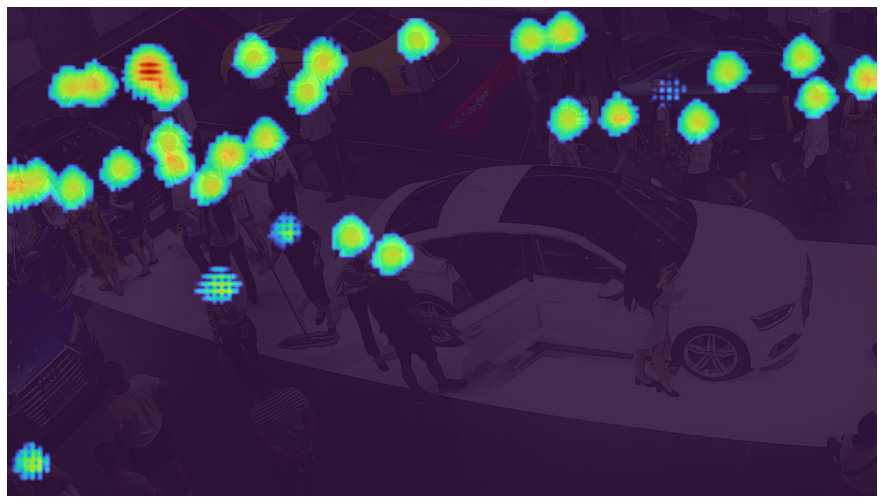}
\label{fig:7.6}
}
\caption{Crowd density predictions of two test images from Crowd Surveillance \cite{yan2019perspective}. The proposed model has accomplished fine results on this dataset, with the test MAE being $4.01$.}
\label{fig:7}
\end{figure}

In practice, $20\%$ training data are dedicated to validation, and similar visualization of predictions is provided in Fig. \ref{fig:7}.

\subsection{Model Comparison} \label{sec:model_comparison}

In this section, the performance of the proposed model on ShanghaiTech datasets \cite{zhang2016single} is compared only with open-source state-of-the-art algirithms, since the computational complexity needs to be measured. Regarding Mall \cite{chen2012feature}, as many approaches that have published results on this dataset have not open-sourced their code, this quantity will not be computed. A randomly generated $1080\times1920$ image is used to calculate the number of arithmetic operations.

\begin{table}[htbp]
\caption{Comparison with state-of-the-art models on ShanghaiTech \cite{zhang2016single}}
\begin{center}
\begin{tabular}{|c|c|c|c|c|c|}
\hline
\multirow{2}{*}{\textbf{Methods}} &
\multirow{2}{*}{\textbf{Complexity$^\mathrm{a}$}} &
\multicolumn{2}{|c|}{\textbf{Part A}} &
\multicolumn{2}{|c|}{\textbf{Part B}} \\
\cline{3-6} 
& & \textbf{\textit{MAE}} & \textbf{\textit{RMSE}}& 
\textbf{\textit{MAE}} & \textbf{\textit{RMSE}}  \\
\hline
MCNN \cite{zhang2016single}  & \textbf{56.21\;G} & 110.2 & 173.2  & 26.4 & 41.3 \\
CMTL \cite{sindagi2017cnn}  & 243.80\;G & 101.3 & 152.4 &  20.0 & 31.1 \\
CSRNet \cite{li2018csrnet}  & 857.84\;G & 68.2 & 115.0 &  10.6 & 16.0 \\
CAN \cite{liu2019context}  & 908.05\;G & 62.3 & 100.0 &  7.8 & 12.2 \\
DM-Count \cite{wang2020distribution} & 853.70\;G & \underline{59.7} & \underline{95.7} & 7.4 & 11.8 \\
M-SegNet \cite{thanasutives2021encoder} & 749.73\;G & 60.55 & 100.8 & \underline{6.8} & \underline{10.4} \\
SASNet \cite{song2021choose} & 1.84\;T & \textbf{54.59} & \textbf{88.38} & \textbf{6.35} & \textbf{9.9} \\
\hline
ICC (proposed) & \underline{125.53\;G} & 76.97 & 130.16 & 8.46 & 15.20 \\
\hline
\multicolumn{6}{l}{
$\mathrm{a}$. The computational complexity of each model is quantified by the
}\\
\multicolumn{6}{l}{
number of arithmetic operations needed to make inference on a $1080$P
}\\
\multicolumn{6}{l}{
image.
}
\end{tabular}
\label{tab:1}
\end{center}
\end{table}

\begin{table}[hbtp]
\caption{Comparison with advanced algorithms on Mall \cite{chen2012feature}}
\begin{center}
\begin{tabular}{|c|c|c|}
\hline
\textbf{Methods} & \textbf{MAE} & \textbf{RMSE} \\
\hline
Chen \textit{et al.} \cite{chen2012feature} & 3.15 & 15.7 \\
ConvLSTM \cite{xiong2017spatiotemporal} & 2.24 & 8.50 \\
DecideNet \cite{liu2018decidenet}       & 1.52 & \underline{1.90} \\
DRSAN \cite{liu2018crowd}               & 1.72 & 2.10 \\
SAAN \cite{hossain2019crowd}            & \textbf{1.28} & \textbf{1.68} \\
LA-Batch \cite{zhou2021locality}        & \underline{1.34} & \underline{1.90} \\
\hline
ICC (proposed) & 2.16 & 2.74 \\
ICC (proposed) + Crowd Surveillance \cite{yan2019perspective} & 3.79 & 4.77 \\
\hline
\end{tabular}
\label{tab:2}
\end{center}
\end{table}

Results are summerized in Table \ref{tab:1} and \ref{tab:2}, showing that the proposed model have comparatively good performances on these three datasets. Compared with the top algorithm SASNet \cite{song2021choose}, the computation can be reduced by $93.18\%$ by sacrificing $41.00\%$, $47.27\%$, $33.23\%$ and $53.54$ performance, under the metrics of MAE and MSE on ShanghaiTech A and B \cite{zhang2016single}, respectively. As for its comparison against the second best models (DM-Count \cite{wang2020distribution} and M-SegNet \cite{thanasutives2021encoder} on ShanghaiTech A and B \cite{zhang2016single} respectively), losses have to increase by $28.93\%$, $36.00\%$, $24.41\%$ and $46.15\%$, in exchange for $85.30\%$ and $83.26\%$ decrease in calculations. More remarkably, on ShanghaiTech B \cite{zhang2016single}, which is more similar to the deployment environment, the proposed architecture outperforms CSRNet \cite{li2018csrnet} under both metrics, with $85.3\%$ operations less, demonstrating its exceedingly high efficiency. 

As for the Mall \cite{chen2012feature} dataset, the proposed method also achieves comparable results on it. Table \ref{tab:2} shows that leading models can control both MSE and RMSE less than $3.0$, and the proposed can also manage to do so. Also, for an ICC that has no access to the training data of Mall \cite{chen2012feature} (denoted as ``ICC (proposed) + Crowd Surveillance \cite{yan2019perspective}'' in Table \ref{tab:2}), exploiting trained weights on Crowd Surveillance \cite{yan2019perspective} is effective, since its tset RMSE is $4.77$, much less than those of Chen \textit{et al.} \cite{chen2012feature} and ConvLSTM \cite{xiong2017spatiotemporal}, and the gap between MAEs of the two ICCs is not immense. This benefit can contribute vastly to the deployment of crowd counting methods. However, because the authors of those approaches listed in Table \ref{tab:2} have not open sourced their code for implementation yet, each model's computational complexity is not compared.

\subsection{Ablation Study} \label{sec:ablation_study}

In this section, the effectiveness of the context-aware module \cite{liu2019context} and those inception blocks \cite{szegedy2016rethinking} is researched by evaluating the performance of ICC without these modules on ShanghaiTech B \cite{zhang2016single}. Except for the structure, everything else in the setting remains the same. Results are shown in Table \ref{tab:3}, showing that including both the contextual module from CAN \cite{liu2019context} and inception blocks \cite{szegedy2016rethinking} can both significantly lower the errors. However, as the difference between the model's performance with and without the contextual module is notably prominent, the contextual module has the major role to take in the extraction of contextual information, and as claimed in previous sections, the function of inception blocks \cite{szegedy2016rethinking} is to reinforce this process.

\begin{table}[hbtp]
\caption{Component Analysis of ICC on ShanghaiTech B \cite{zhang2016single}}
\begin{center}
\begin{tabular}{|c|c|c|}
\hline
\textbf{Methods} & \textbf{MAE} & \textbf{RMSE} \\
\hline
Original & 8.46 & 15.20 \\
No Contextual Module & 25.56 & 44.72 \\
No Inception Blocks & 10.62 & 17.83 \\
\hline
\end{tabular}
\label{tab:3}
\end{center}
\end{table}

\section{Conclusion}

In this paper, a convolutional neural network, established on Inception-V3 \cite{szegedy2016rethinking} and CAN \cite{liu2019context} has been proposed to facilitate the deployment of crowd counting models in surveillance systems. The proposed method is much less computationally complex compared with the state-of-the-art algorithms, while its performance is not significantly harmed. This property has been testified by various experiments on benchmark datasets. Both context-aware components within the model have been proved to be useful, and this work also shows models pre-trained on Crowd Surveillance \cite{yan2019perspective} have good generalization.

\section*{Acknowledgment}

I would like to present my gratitude to Prof. Tanaya Guha and Prof. Victor Sanchez. This work was complete under their supervision and guidance. Also, special thanks go to the owner (Junyu Gao) and contributors of the great GitHub repository \href{https://github.com/gjy3035/Awesome-Crowd-Counting}{Awesome Crowd Counting}, in which comprehensive information about crowd counting datasets and benchmarks are provided.

\small{\printbibliography}

\end{document}